# PointMoment: Mixed-Moment-based Self-Supervised Representation Learning for 3D Point Clouds

Xin Cao[1,2], Xinxin Han[1,2], Yifan Wang[1,2], Mengna Yang[1,2], Kang Li[1,2,3]

[1]School of Information Science and Technology, Northwest University, Xi'an, Shaanxi 710127, China
[2]National and Local Joint Engineering Research Center for Cultural Heritage Digitization, Xi'an, Shaanxi 710127, China

[3]likang@nwu.edu.cn

**Abstract:** Large and rich data is a prerequisite for effective training of deep neural networks. However, the irregularity of point cloud data makes manual annotation time-consuming and laborious. Self-supervised representation learning, which leverages the intrinsic structure of large-scale unlabelled data to learn meaningful feature representations, has attracted increasing attention in the field of point cloud research. However, self-supervised representation learning often suffers from model collapse, resulting in reduced information and diversity of the learned representation, and consequently degrading the performance of downstream tasks. To address this problem, we propose PointMoment, a novel framework for point cloud self-supervised representation learning that utilizes a high-order mixed moment loss function rather than the conventional contrastive loss function. Moreover, our framework does not require any special techniques such as asymmetric network architectures, gradient stopping, etc. Specifically, we calculate the high-order mixed moment of the feature variables and force them to decompose into products of their individual moment, thereby making multiple variables more independent and minimizing the feature redundancy. We also incorporate a contrastive learning approach to maximize the feature invariance under different data augmentations of the same point cloud. Experimental results show that our approach outperforms previous unsupervised learning methods on the downstream task of 3D point cloud classification and segmentation.

## 1. Introduction

With the rapid development of virtual reality, metaverse and other technologies, there is an increasing demand for accurate capture and simulation of the real world[1]. Point cloud, a data format that effectively represents the 3D world, can be directly acquired by entry-level depth sensors and outperforms triangular meshes or voxels in processing and storage efficiency[2]. Therefore, point clouds are widely applied to various 3D scene understanding tasks, such as object segmentation[3] and classification[4]. In recent years, deep learning has become a dominant paradigm for point cloud understanding. Many innovative deep learning methods for point clouds have been proposed, such as PointNet[5], PointNet++[6], DGCNN[7], and so on. However, these deep learning methods rely heavily on large-scale annotated data, which is difficult to obtain for point cloud data due to its sparse, irregular and low-resolution nature. Self-supervised representation learning, which has shown great potential in domains such as computer vision and natural language processing, can offer new insights for addressing the above issues[8,9].

Self-supervised representation learning aims to learn robust and general feature representations from unlabeled data to improve the performance of various downstream tasks.[10] In recent years, several self-supervised representation learning algorithms have been developed for point clouds, such as point cloud self-reconstruction[11,12], point cloud adversarial generation[13,14], and point cloud completion[15,16]. Among them, Latent GAN[13] requires additional computing resources and time to train the autoencoder and a generative adversarial network, and FoldingNet[11], OcCo[16] also face the same problem. Anyway, these algorithms suffer from high sensitivity to rotation and translation variations. Inspired by the success of contrastive learning in image and video domains[17,18], more and more studies have attempted to apply it to point cloud understanding[19–21]. PointContrast[19] employs contrastive learning to learn viewpoint-invariant representations of point clouds, which enable high-level scene understanding tasks. CrossPoint[22] further leverages different modalities for contrastive learning and extracts rich learning signals from them. These methods perform pre-training by designing a pretext task that pulls similar samples closer and pushes dissimilar samples farther in the projected space. However, model collapse is a common problem in self-supervised representation learning, which leads to the shrinkage of the learned representation vector to a constant or low-dimensional subspace, and thus fails to fully utilize the entire representation space. Most existing contrast-based learning methods address this problem by employing complex and fine-tuned mechanisms, such as memory banks[20] or asymmetric networks(gradient stopping, predictor network, momentum update strategies etc.)[21]. Therefore, how to prevent model collapse without relying on complex designs is still an open challenge.

In this paper, we propose a self-supervised contrastive learning method for point clouds based on high-order mixed moment, which effectively prevents model collapse by minimizing the redundancy among feature variables, without relying on the complex mechanisms mentioned above. Some methods in 2D images introduce a covariance matrix to minimize the redundancy between pairs of variables by forcing them closer to the identity matrix[23,24]. However, these methods fail to reduce the redundancy among multiple variables. HOME[25] addresses this limitation and can eliminate redundant information among multiple variables. Inspired by this, we incorporate the idea of feature redundancy into point cloud self-supervised representation learning, and leverage high-order mixed moment to minimize the redundancy among arbitrary feature variables for learning meaningful point cloud representations. It is known that pairwise independence of each variable does not guarantee their mutual independence, and that the total correlation among all variables is minimized only if multiple variables are independent. The independence of the variables implies that the mixed moment of multiple feature variables can be decomposed into the product of their individual moment. Based on this statistical theory and the contrastive learning paradigm, we design a loss function based on high-order mixed moment to reduce the redundancy among multiple variables, enabling the high-dimensional embedding features learned by self-supervised representation learning to contain rich and independent information and ensuring the maximum representation consistency of different augmented point clouds. We evaluate our method on several datasets through extensive experiments, which demonstrate that our approach achieves state-of-the-art accuracy.

The main contributions of this paper are summarized as follows:
(1) We propose a novel self-supervised representation learning method for point clouds based on high-order mixed moments, which enhances the representation capability of the model by minimizing the redundancy among feature variables.

(2) We apply the high-order mixed moment to point cloud self-supervised representation learning for the first time, offering new insights for future self-supervised learning of point clouds.

(3) We evaluate our approach on two common downstream tasks, namely, object classification and segmentation, using various synthetic and real-world datasets, and demonstrate that PointMoment surpasses previous unsupervised learning methods.

## 2. Related works

In this section, we briefly review the recent research progress on two related topics: supervised point cloud representation learning, and self-supervised point cloud representation learning.

### 2.1 Supervised representation learning on point clouds

We can categorize supervised representation learning methods for point clouds into two types: structure-based and point-based methods. Structure-based approaches typically project point clouds into 2D images or convert them into regular structured data such as voxels for feature learning. MVCNN[26] applies 2D convolutional networks and max-pooling to aggregate multi-view features into compact global shape descriptors. However, this approach only focuses on the max-feature elements, leading to information loss and neglect of view relationships. View-GCN[27] addresses this limitation by employing a GCN on the view graph to hierarchically aggregate multi-view features. VoxNet[28] extracts features from voxel grids using 3D convolutional networks. However, these methods face memory and computational challenges when handling dense 3D data, which grows cubically with resolution. In summary, these methods tend to fail to capture fine-grained geometric information and consume a lot of memory.

Point-based methods directly utilize raw point cloud data without transforming them into other formats. We can categorize these methods into four types: MLP-based, CNN-based, Graph-based, and Transformer-based. The MLP-based method extracts features from each point in the point cloud data. PointNet[5], a pioneering work in this field, inspires many subsequent methods, such as PointWeb[29]. However, PointNet fails to extract local information from point clouds. PointNet++[6] addresses this limitation by using set abstraction structures to extract local features and generate compact global representations through aggregation. PointMLP[30], on the other hand, focuses less on local geometric information and uses a simple feedforward residual MLP network for point cloud representation learning. CNN-based methods extract point cloud features by designing convolution kernels. PointCNN[31] introduces the X-Conv operator that enables traditional convolution to directly process point cloud data. SpiderCNN[32] uses step functions and third-order Taylor expansion products to capture complex local geometric information. PAConv[33] takes a data-driven approach for adaptive convolution kernels to handle disordered point cloud data better. Graph-based methods treat points as vertices of a graph and build directed edges based on their relationships with neighboring points for feature learning. DGCNN[7] reconstructs the local graph using the K-nearest neighbor algorithm after each layer of graph convolution and introduces the EdgeConv representation module for extracting local geometric features. However, it captures limited edge features. AdaptConv[2] generates adaptive kernels based on the dynamically learned point features, enhancing its ability to capture relationships between points during the graph convolution process. Nonetheless, these methods still face challenges in exploring multi-scale information. Recently, self-attention mechanisms have achieved significant success in the field of NLP and images. Some studies have begun to explore their application to point cloud data, such as PCT[34] and PointTransformer[35].

In summary, the lack of large annotated datasets is the major challenge for such supervised learning approaches.

## 2.2 Self-supervised representation learning on point clouds

Self-supervised representation learning, which can extract effective features without annotated data, emerges as a promising research direction in point cloud learning. We can divide them into two types: generative and contrastive methods.

Generative methods usually attempt to use generative models to model the latent distribution of data or features to learn the hidden features of point cloud data. Some common generative models are variational autoencoders, generative adversarial networks, etc. Latent-GAN[13] is a pioneer in point cloud generation models, which applies GAN to both raw point cloud data and embedding features and produces high-quality point cloud samples. Yang et al.[11] assumed that 3D object surfaces are essentially 2D manifolds that can be mapped from 2D planes through folding operations. They designed a folding-based decoder that deforms regular 2D meshes to 3D object surfaces, improving the reconstruction quality. With the widespread use of transformers in image and natural language processing, many transformer-based generative methods have emerged. Point-BERT[12] trains transformers by predicting masked tokens of point clouds, enhancing the performance of transformers on point clouds. Point-MAE[36] is an extension of Point-BERT. Recently, Wang et al. introduced OcCo[16], a point cloud completion method that can learn representations by reconstructing the missing parts of incomplete point cloud data. However, these methods often fail to reconstruct point cloud data accurately and consume a lot of computing resources.

Contrastive learning methods train a feature encoder to learn representations that are similar to positive samples and dissimilar to negative samples. Info3D[20] combines mutual information and contrastive learning for the first time to improve the model's ability to represent 3D objects by maximizing the mutual information between the global shape and the local structures of 3D objects. PointGLR[37] integrates contrastive learning, self-reconstruction, and normal estimation into a unified framework and learns bidirectional inference between the local structure and the global shape of 3D objects in an unsupervised manner. Recently, PointContrast[19] unified the contrastive learning framework for 3D point cloud representation learning and promotes high-level scene understanding through contrastive learning on different views of the point cloud. STRL[21] extends BYOL[38] briefly and applies it to point cloud representation learning. DepthContrast[39] proposes a joint contrastive learning method for point clouds and voxels to improve the model performance on downstream tasks further. CrossPoint[22] introduces 2D images as rich learning signals for 3D point cloud understanding and proposes a simple yet powerful cross-modal contrastive learning method. However, obtaining 2D-rendered images of point clouds can be difficult, which limits the applicability of this method.

In contrast to these methods that often rely on complex mechanisms to prevent model collapse (e.g., Info3D's memory bank, STRL's asymmetric network, etc.), we propose a simple contrastive learning network framework with a novel loss function that achieves an ideal representation of point clouds.

## 3. Methodology

In this section, we present PointMoment, a novel self-supervised representation learning approach that leverages high-order mixed moment to learn meaningful representations. Firstly, we introduce the concept of high-order mixed moment(Sec 3.1) and subsequently present a loss function(Sec 3.2) that pre-trains to the feature extractor $f_\theta(\cdot)$ under a self-supervised framework. Fig. 1 illustrates the network framework of our approach. Lastly, we showcase an application of our approach to self-supervised representation learning based on the third-order mixed moment(Sec 3.3), which is illustrated in Fig. 2.

## 3.1 High-order mixed-moment

Moment are numerical characteristics that describe the distribution of random variables in probability theory and statistics. The most commonly used moment are first-order moment (EX) and second-order central moment (DX), which measure the average level and dispersion of values of a single random variable, respectively. To obtain a more comprehensive measurement of multiple random variables, mixed moment are introduced.

Mixed moment are statistics that describe the basic characteristics of the distribution of a multivariate random variable. They are usually defined as follows:

For any positive integer $k_i$, the mathematical expectation $E[X_1^{k_1} \ldots X_n^{k_n}]$ of multiple random variables is called the k-order mixed moment, where $k = k_1 + \ldots k_n$. $E[(X_1 - E[X_1])^{k_1} \ldots (X_n - E[X_n])^{k_n}]$ is then called the k-order central mixed moment. If there are only two random variables and both $k_i$ are 1, then the mixed moment $E[(X_1 - E[X_1])E(X_2 - E[X_2])]$ is called the covariance, which describes the degree of correlation between the two random variables. Zbontar et al.[23] just start from the perspective of covariance and reduce the feature redundancy by reducing the correlation between any two feature variables.

After discussing the concept of mixed moment, we can now explain the motivation for using them. In high-dimensional feature spaces, embedding features often contain redundant information that can hinder their effectiveness. To address this problem, we adopt Shannon's concept of mutual information[40] to quantify the interdependence between random variables. Let $X_1, X_2, \ldots X_d$ be the random variables representing each dimension (denoted as d) of the embedding feature. We can use the mutual information (MI) formula to measure the shared information between variables:

$$I(X_1, X_2, \ldots, X_d) = \int_{X_1} \int_{X_2} \ldots \int_{X_d} p(x_1, x_2, \ldots, x_d) \log \frac{p(x_1, x_2, \ldots, x_d)}{p(x_1)p(x_2)\ldots p(x_d)} dx_1 dx_2 \ldots dx_d \quad (1)$$

where $p(x_1, x_2, \ldots, x_d)$ is the joint density function of $X_1, X_2, \ldots X_d$ and $p(x_1)$, $p(x_2)$, ..., $p(x_d)$ is the marginal probability density function of $X_1, X_2, \ldots X_d$ respectively.

To make the information contained in the embedding features rich and compact, we need to minimize Eq.1 to reduce the redundant information between feature variables. According to statistics, Eq.1 can be minimized when $X_1, X_2, \ldots X_d$ are independent of each other, that is, $p(x_1, x_2, \ldots, x_d) = p(x_1)p(x_2)\ldots p(x_d)$. However, this minimization process faces two key challenges. One challenge is that pairwise independence of random variables does not necessarily ensure that the total redundancy of all variables is minimized, unless $p(x_1, x_2, \ldots, x_d)$ follows a multivariate normal distribution, which is often hard to satisfy in practice. Another challenge is that directly modelling the probability distribution of continuous variables can be difficult. Therefore, a common approach is to use statistical moment to model the probability distribution. We introduce high-order mixed moment. We exploit the fact that all random variables are independent of each other if and only if the expectation of all random variables is equal to the product of their individual expectations, which can be mathematically expressed as follows:

$$E[\prod_{d=1}^{D} X_d] = \prod_{d=1}^{D} E[X_d] \quad (2)$$

where $E[\prod_{d=1}^{D} X_d]$ is the mixed moment of order D. We can minimize Eq.1 by modelling and enforcing the random variables to satisfy the condition of Eq.2, which induces more independence among them.

## 3.2 Formulation of high-order mixed-moment as a loss

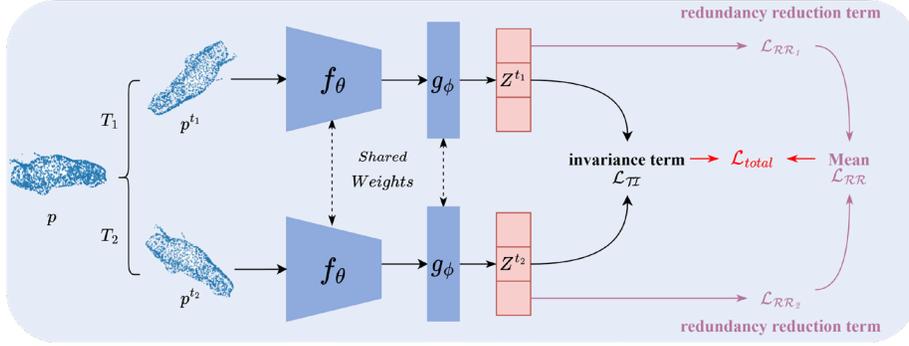

Fig. 1. The overall architecture of our approach PointMoment. The loss is composed of two important parts, one is the loss based on invariance, and the other is the loss based on redundancy reduction. The latter involves constraints of order two or higher.

### 3.2.1 Preliminary

Given a batch of randomly selected point clouds $B = \{p_i\}_{i=1}^{|B|}$, where $p_i \in R^{N \times 3}$, $|B|$ denotes batch size and N denotes the total number of points in each point cloud. We apply random data augmentation techniques such as rotation, scaling, and jittering to each point cloud $p_i$, generating two distinct enhanced versions of the original data, $p_i^{t_1}$ and $p_i^{t_2}$. Next, we feed both enhanced datasets into a network architecture. Specifically, the feature extractor $f_\theta$ maps $p_i^{t_1}$ and $p_i^{t_2}$ into a feature representation space, followed by a projection head $g_\phi$, which projects resulting feature vectors onto an embedding feature space. We denote the final embedding vectors produced from this process as $z_i^{t_1}$ and $z_i^{t_2}$, respectively, where $z_i^t = g_\varphi(f(p_i^t; \theta_f); \varphi_g)$ and $z_i^t \in R^D$, $\theta_f$ and $\varphi_g$ denote the trainable parameters of the encoder and projector, respectively, and D denotes the dimension of the embedding vector. Our goal is to learn compact, meaningful feature vectors without the need for complex network architectures or optimization processes. Furthermore, our loss function can be used in combination with a range of other network frameworks. Fig. 1 provides a general representation of our self-supervised learning approach for point clouds.

### 3.2.2 Loss based on high-order mixed-moment

We argue that a desirable embedding feature must have two essential properties: invariance to randomly augmented samples and minimized total correlation between all variables of the embedded vector. Accordingly, we design separate loss functions for the above two properties.

**Redundancy reduction based on high-order mixed-moment:** Reducing redundancy encourages the network to learn a compact embedding vector in which each variable can express distinct semantic information. As discussed in Sec 3.1, we can enforce Eq.2 to achieve the minimum total correlation among all variables to some degree. We use Shinchin's Law of Large Numbers [41] to approximate the expectation of a variable by sample moment, which are then applied to compute the expectation in Eq.2. For simplicity, we standardize each feature variable z along the batch dimension to have zero mean and unit variance, as follows:

$$\hat{z}_{b,d}^t = \frac{z_{b,d}^t - \frac{1}{B}\sum_{b=1}^{B} z_{b,d}^t}{\sqrt{\sum_{b=1}^{B}(z_{b,d}^t - \frac{1}{B}\sum_{b=1}^{B} z_{b,d}^t)^2 / B}} \quad (3)$$

Now for each dimension satisfies $E[\hat{Z}_d] = 0$, so we only need to set $E[\prod_{d=1}^{D} \hat{Z}_d] = 0$ to make Eq.2 hold, that is, $E[\prod_{d=1}^{D} \hat{Z}_d] = \prod_{d=1}^{D} E[\hat{Z}_d] = 0$. Therefore, we propose the redundancy reduction loss based on the high-order mixed moment as follows:

$$\mathcal{L}_{RR} = \frac{1}{T}\sum_{t=1}^{T}\left[\frac{1}{2M}\sum_{K=2}^{D}\left(\sum_{d_1}^{D}\sum_{d_2\neq d_1}^{D}\cdots\sum_{d_K\neq\cdots\neq d_2\neq d_1}^{D}\left(E_{d_1,d_2,\cdots,d_K}\right)^2\right)\right] \quad (4)$$

where K denotes the order of the mixed moment and M denotes the total number of combinations of mixed moment of all orders, i.e. $M = \sum_{K=2}^{D}\frac{D!}{(D-K)!K!}$. $E_{d_1,d_2,\cdots,d_K}$ is the matrix of mixed moment of order K calculated along the batch dimension:

$$E_{d_1,d_2,\cdots,d_K} = \frac{1}{B}\sum_{b=1}^{B}\prod_{i=1}^{K}\hat{z}_{b,d_i}^{t} \quad (5)$$

where $d_1, d_2, \cdots, d_K$ etc. denote the index of the dimension of the embedding vector, respectively, for any K variables with $K \leq D$ and with $1 \leq d_K \leq D$ for any $d_K$. $E_{d_1,d_2,\cdots,d_K}$ denotes the value of the K-order mixed moment of the random variables with index value $d_1, d_2, \cdots, d_K$, respectively.

**Transformation invariance based on co-correlation:** Invariance refers to the property that semantically similar point cloud data are mapped to nearby regions in the embedding feature space. Many existing methods compute cosine similarity between two different augmented versions of the point cloud and design contrastive losses based on it. In contrast, we adopt the co-correlation matrix proposed by Zbontar et al.[23] to enhance invariance and propose the following loss function:

$$\mathcal{L}_{TI} = \frac{1}{D}\sum_{d=1}^{D}(1 - C_{dd})^2 \quad (6)$$

where C is the co-correlation matrix, calculated from the outputs of two identical networks along the batch dimension:

$$C_{i,j} = \frac{\sum_{b=1}^{B}\hat{z}_{b,i}^{t_1}\hat{z}_{b,j}^{t_2}}{\sqrt{\sum_{b=1}^{B}\left(\hat{z}_{b,i}^{t_1}\right)^2}\sqrt{\sum_{b=1}^{B}\left(\hat{z}_{b,j}^{t_2}\right)^2}} = \frac{1}{B}\sum_{b=1}^{B}\hat{z}_{b,i}^{t_1}\hat{z}_{b,j}^{t_2} \quad (7)$$

where i and j denote the dimensional index of the embedding vector, C is a square matrix of size D and $-1 \leq C_{i,j} \leq 1$, -1 denotes negative correlation and 1 denotes positive correlation. It is analytically easy to see that Eq.7 can eventually be reduced to the same expression as Eq.5, which essentially satisfies the same underlying logic. Eq.6 maximizes the correlation of the different augmented embedding features by forcing the diagonal elements of the co-correlation matrix to be 1, thus satisfying the transformation invariance.

Based on the above analysis, we give the final loss:

$$\mathcal{L}_{total} = \mathcal{L}_{TI} + \lambda\mathcal{L}_{RR} \quad (8)$$

where λ is a parameter used to trade off the first and second terms.

*3.3 Instantiating three-order mixed-moment for contrastive learning*

Based on Sec 3.1 and Sec3.2, we instantiate a point cloud self-supervised representation learning approach that leverages third-order mixed moment, as shown in Fig. 2(a). We also present a more efficient version in Fig. 2(b), which relaxes the constraint on all branches of the network. Instead, it randomly chooses any branch to produce the same effect, thus reducing the computational cost. We present experimental results to demonstrate this approach in the following section.

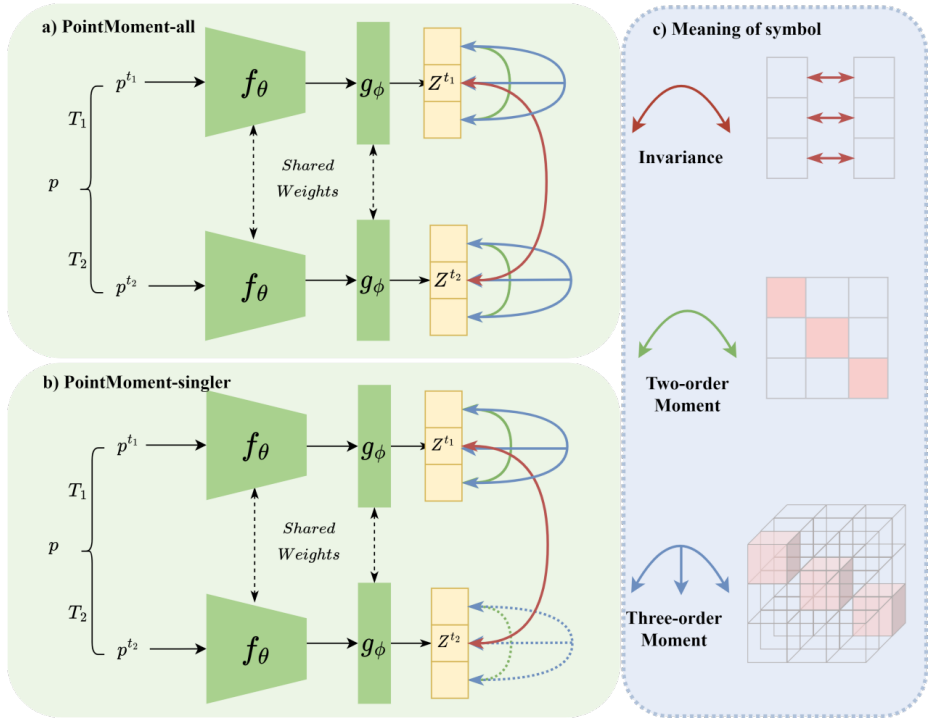

Fig. 2. Self-supervised representation learning based on third-order mixed moment. a) indicates that mixed moment are calculated for the output of each branch, and b) indicates that mixed moment are calculated for the output of a particular branch, where the dashed line indicates that the corresponding output is not constrained by the mixed moment. c) visualization shows the specific meaning of each curve, and the red square part does not take into account the calculation of the mixed moment.

Fig. 2 shows that our approach based on third-order mixed moment calculates the total loss $\mathcal{L}_{\text{total}}$ when K={2,3}. The green curve represents the second-order mixed moment calculated for the embedding features of the branch output. This calculation degenerates into Ztonbar et al.'s [23] approach at this point. Similarly, the blue curve represents the third-order mixed moment computed for the branch output. These two constraints are combined, i.e. $\mathcal{L}_{\text{RR}}$, to eliminate redundancy in the embedding features. Finally, the red curve signifies the consistency of data after undergoing two transformations, i.e. $\mathcal{L}_{\text{TI}}$, which completes the transformation invariance constraint.

## 4. Experiments and results

In this section, we first introduce the pre-training setup and the datasets used. Next, we evaluate the transferability of our approach on two common downstream tasks, namely, object classification and segmentation. Finally, we conduct ablation experiments to examine the effectiveness of our loss function and parameter settings.

### 4.1 Pre-training

**Dataset.** We used the ModelNet40 dataset to pre-train PointMoment, which consists of CAD models of 40 common objects. We use the same approach as STRL[21] to obtain 12311 point cloud data by randomly sampling 2048 points from these CAD models, of which 9843 are used for training and 2468 for testing. Additionally, we adopted a range of standard data augmentation techniques, including geometric transformations such as rotation, translation and scaling, and spatial transformations such as jittering and normalization. We applied these transformations randomly to generate an augmented pair of point clouds for each input, and fed them into the network for pre-training.

**Implementation details.** To fairly compare our approach with existing techniques, we use PointNet[5] and DGCNN[7] as point cloud feature extractors and a multilayer perceptron containing two layers as the projection head, which maps the feature vector into the embedding space to generate a 512-dimensional embedding vector. We trained the model in an end-to-end manner for 200 rounds using an Adam optimizer with a weight decay of $1\times10^{-6}$ and an initial learning rate of $1\times10^{-3}$, while adjusting the learning rate using a learning rate decay strategy with cosine annealing. After pre-training, we discarded the projection head $g_\phi(\cdot)$ and retained $f_\theta(\cdot)$ for the following downstream task.

*4.2 Downstream tasks*

4.2.1 3D Object classification

The 3D object classification task aims to classify the given point cloud data and accurately predict the specific object class to which each point cloud belongs. We evaluate the shape understanding ability and the generalization performance of pre-trained models using two widely-used benchmark datasets: ModelNet40 and ScanObjectNN[42]. ModelNet40 is a synthetic object dataset consisting of 40 categories and 12,311 CAD models, which we use to evaluate the classification performance on synthetic objects. ScanObjectNN is a highly-challenging, realistic 3D point cloud dataset collected from real indoor scenes, which we use to evaluate the performance in real natural scenes. The dataset contains 15 categories with a total of 2880 objects, of which 2304 are used for training and 576 for testing.

Our PointMoment can be flexibly combined with different backbone networks. We chose two widely used backbone networks, namely PointNet and DGCNN, as point cloud feature extractors, and following the standard protocol of previous work, we conducted a comprehensive evaluation of the performance of our method on object classification. Specifically, we used a simple linear support vector machine (SVM) classifier that was trained on the 3D features learned from the training set by the point cloud feature extractor with the parameters frozen, and we applied this classifier to the 3D features of the test set to obtain the classification results.

**Synthetic object classification.** Table 1 shows the accuracy of PointMoment for linear classification on ModleNet40. To reduce the consumption of computational resources, we randomly selected one of the network branches for high-order mixed moment constraint, which produced essentially the same results as constraining all network branches. We used single-branch constraints for comparison in the subsequent experiments.

When using PointNet or DGCNN as the backbone network, our approach outperforms other state-of-the-art unsupervised or self-supervised algorithms. We note that our approach employs only the basic network architecture without STRL's complex asymmetric network, gradient stopping or other designs. In both cases of using PointNet or DGCNN as the backbone network, our approach outperforms STRL by 0.5% and 0.1% respectively, demonstrating the effectiveness of our approach.

**Real-world object classification.** We evaluated the generalizability of PointMoment in real-world scenarios by testing ScanObjectNN using SVM. Table 2 shows the linear classification results of different self-supervised methods on ScanObjectNN. Compared with the previous state-of-the-art methods, our approach achieved 4.8% and 2.6% higher accuracy than SOTA using PointNet and DGCNN as the backbone network, respectively. This outcome underscores the generalizability of representations learned on synthetic datasets, verifying the effectiveness of our approach.

Table 1. Comparison of the linear SVM classification on ModelNet40. The linear classifier is fitted on the training set of ModelNet40 using the pre-trained model, and the model performance is evaluated on the test set.

| Method | ModelNet40 |
| --- | --- |
| 3D-GAN[14] | 83.3 |
| Latent-GAN[13] | 85.7 |
| SO-Net[43] | 87.3 |
| FoldingNet[11] | 88.4 |
| MRTNet[44] | 86.4 |
| 3D-PointCapsNet[45] | 88.9 |
| MAP-VAE[46] | 88.4 |
| DepthContrast[39] | 85.4 |
| Jigsaw[47]+PointNet | 87.3 |
| Rotation[48]+PointNet | 88.6 |
| OcCo[16]+PointNet | 88.7 |
| STRL[21]+PointNet | 88.3 |
| **PointMoment-all(Ours)+PointNet** | **88.8** |
| **PointMoment-singler(Ours)+PointNet** | **88.8** |
| Jigsaw[47]+DGCNN | 90.6 |
| Rotation[48]+DGCNN | 90.8 |
| STRL[21]+DGCNN | 90.9 |
| OcCo[16]+DGCNN | 89.2 |
| PointMoment-all(Ours)+DGCNN | 90.9 |
| **PointMoment-singler(Ours)+DGCNN** | **91.0** |

Table 2. Comparison of classification on ScanObjectNN. PointMoment achieves improvements compared to other self-supervised methods on both PointNet and DGCNN, which illustrates the effectiveness of our method in real-world scene classification.

| Encoder | Method | Acc. |
| --- | --- | --- |
| PointNet | Jigsaw[47] | 55.2 |
|  | OcCo[16] | 69.5 |
|  | STRL[21] | 74.2 |
|  | **PointMoment(Ours)** | **79.0** |
| DGCNN | Jigsaw[47] | 59.5 |
|  | OcCo[16] | 78.3 |
|  | STRL[21] | 77.9 |
|  | **PointMoment(Ours)** | **80.5** |

### 4.2.2 3D Object part segmention

Object part segmentation is an important and challenging 3D recognition task, where the goal is to assign a part class label to each point, such as a table leg or a car tire. We conducted experiments on the ShapeNetPart[49] dataset, which contains 16991 objects from 16 categories, with a total of 50 parts distributed among 2 to 6 parts per object. As a widely used benchmark dataset, ShapeNetPart can effectively evaluate the performance of object part segmentation methods. We used DGCNN as the backbone network in the pre-training phase and applied fine-tuning to optimize the performance of our method. To evaluate the performance of our method, we used the mean intersection over union (mIoU) metric, which is highly accurate and commonly used.

Table 3 summarizes the segmentation results of supervised learning methods and different self-supervised methods on ShapeNetPart. Compared to DGCNN parameters randomly initialized by supervised learning, our pre-training approach provided superior initial weights for DGCNN, boosting mIoU by 0.3%. Our model also surpassed the current state-of-the-art self-supervised method by 0.3% in mIoU, which demonstrates that our method uses high-order mixed moment to obtain more discriminative and less redundant features.

Table 3. Part segmentation results on ShapeNetPart dataset. Our method outperforms supervised learning methods with random initial weights and other self-supervised learning methods with pre-trained weights.

| Category | Method | mIoU |
|---|---|---|
| Supervised | PointNet[5] | 83.7 |
|  | PointNet++[6] | 85.1 |
|  | **DGCNN[7]** | **85.1** |
| Self-Supervised | PointContrast[19] | 85.1 |
|  | Jigsaw[47] | 84.3 |
|  | OcCo[16] | 85.0 |
|  | **PointMoment(Ours)** | **85.4** |

### 4.3 Ablations and analysis

**Impact of high-order mixed moment.** To evaluate the effectiveness of high-order mixed moment, we designed ablation experiments using three loss functions: i) an invariance-based loss function as a baseline; ii) adding second-order mixed moment loss to i) to assess the effectiveness of the second-order mixed moment; and iii) appending high-order mixed moment loss to ii) to comprehend intuitively the effect of the third-order mixed moment on the results. The related comparative results are shown in Table 4. Using solely the invariance-based loss function caused uneven feature distribution and significantly lowered classification accuracy. Incorporating second-order mixed moment avoided model collapse and reduced feature redundancy, significantly enhancing the representation ability of the model for both PointNet and DGCNN on both datasets. The introduction of third-order mixed moment further decreased feature redundancy. Compared to solely incorporating second-order mixed moment, the inclusion of third-order mixed moment improved classification accuracy by 0.8% and 1.7% for PointNet and DGCNN on the ModelNet40 dataset, respectively. Our experiment demonstrates that it is crucial to introduce higher-order mixed moment.

Table 4. The accuracy of linear SVM classification using retrained embedding on ModelNet40 and ScanObjectNN for PointMoment.

| Encoder | invariance | two-order mixed moment | three-order mixed moment | Acc. ModelNet40 | ScanObjectNN |
|---|---|---|---|---|---|
| PointNet | √ |  |  | 40.5 | 40.6 |
|  | √ | √ |  | 88.0 | 73.4 |
|  | √ | √ | √ | **88.8** | **75.4** |
| DGCNN | √ |  |  | 77.3 | 56.6 |
|  | √ | √ |  | 89.3 | 79.3 |
|  | √ | √ | √ | **91.0** | **80.5** |

Fig. 3 illustrates the visualization results of features obtained on the ModelNet10 test set using t-SNE, with features extracted from the pre-trained PointNet. The introduction of second-order mixed moment and third-order mixed moment enhanced the distinctiveness of different classes, especially after the addition of the third-order mixed moment, we can better distinguish some objects with less clear boundaries (such as sofas, beds and bathtubs).

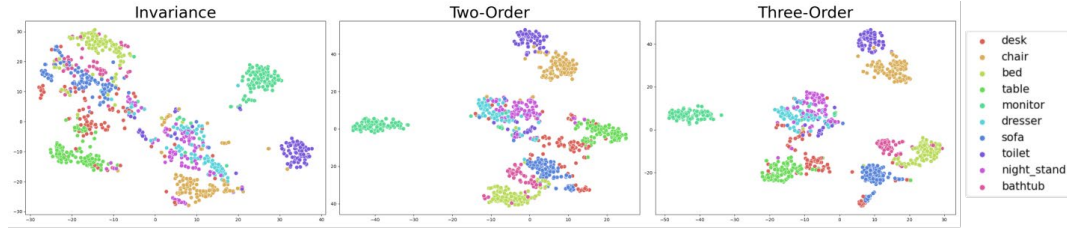

Fig. 3. T-SNE visualization of features on the test split of ModelNet10 after training the PointNet as the backbone in a self-supervised manner. The features learned by three-order mixed moment(right) provide better discrimination of classes(e.g., sofas, beds and bathtubs) than using only invariance(left) or two-order mixed moment(middle).

**Sensitivity analysis of λ.** Intuitively, λ has a crucial impact on pre-training, which in turn affects the performance of downstream tasks. Therefore, we explore the impact of different values of λ on downstream classification tasks. We vary λ from 0.001 to 5 and used PointNet as the backbone network for pre-training. Finally, we performed linear classification on ModelNet40 and ScanObjectNN datasets respectively. From the results in Table 5, when λ is set to 0.5, the classification results achieve the best performance on both datasets.

Table 5. Linear classification results for different λ parameters on ModelNet40 and ScanObjectNN datasets after pre-training using PointNet.

| λ | Acc. | |
|---|---|---|
| | ModelNet40 | ScanObjectNN |
| 0.001 | 72.5 | 57.3 |
| 0.005 | 78.5 | 62.9 |
| 0.01 | 81.2 | 66.6 |
| 0.05 | 84.4 | 71.4 |
| 0.1 | 88.5 | 74.8 |
| **0.5** | **88.8** | **79.0** |
| 1 | 88.0 | 76.4 |
| 5 | 87.8 | 74.1 |

## 5. Conclusion

In this paper, we propose PointMoment, a novel self-supervised representation learning method for point clouds based on high-order mixed-moment. Our method leverages the statistical properties of high-order mixed-moment to reduce the correlation among feature dimensions in the latent space, leading to informative and discriminative embedding features. Unlike existing self-supervised methods for point clouds, our method does not rely on auxiliary components such as predictors or momentum encoders, and prevents model collapse by using a simple symmetric network architecture. We demonstrate the effectiveness and transferability of our learned representations on downstream tasks such as classification and segmentation, even though our method is only pre-trained on a synthetic 3D object dataset. Moreover, our method can be easily integrated with existing contrastive learning frameworks. One limitation of our method is its high memory consumption, which we plan to address in the future by improving the computation of high-order mixed-moment. Besides, we also plan to explore higher-order mixed-moment for learning more useful representations without increasing memory consumption, and further enhance the performance of our method. In conclusion, our work demonstrates the great potential of a novel technique based on higher-order mixed moment and aims to provide new research insights for the field of point cloud representation learning.

**Funding.** National Key Research and Development Program of China (2019YFC1521102, 2019YFC1521103); Key Research and Development Program of Shaanxi Province (2019GY215, 2021ZDLSF06-04); National Natural Science Foundation of China (61701403,61806164); China Postdoctoral Science Foundation (2018M643717).

**Declaration of Competing Interest.** The authors declare no conflicts of interest.

**Data availability.** Data underlying the results presented in this paper can be obtained from the internet.

**Acknowledgment.** We would like to acknowledge Dr. Chuang Niu from Rensselaer Polytechnic Institute for providing technical guidance and expertise that greatly assisted our research.